# FOXANN: A Method for Boosting Neural Network Performance

**Mahmood A. Jumaah[1]***   **Yossra H. Ali[2]**   **Tarik A. Rashid[3]**   **S. Vimal[4]**

[1,2]*Department of Computer Science, University of Technology- Iraq, Baghdad, Iraq*
[3]*Department of Computer Science and Engineering, University of Kurdistan Hewlêr, Erbil, Iraq*
[4]*Department of Artificial Intelligence and Data Science, Ramco Institute of Technology, Rajapalayam, India*

\* Corresponding author's Email: cs.22.27@grad.uotechnology.edu.iq
[2] Yossra.H.Ali@uotechnology.edu.iq
[3] tarik.ahmed@ukh.edu.krd
[4] vimal@ritrjpm.ac.in

**Abstract**
Artificial neural networks play a crucial role in machine learning and there is a need to improve their performance. This paper presents FOXANN, a novel classification model that combines the recently developed Fox optimizer with ANN to solve ML problems. Fox optimizer replaces the backpropagation algorithm in ANN; optimizes synaptic weights; and achieves high classification accuracy with a minimum loss, improved model generalization, and interpretability. The performance of FOXANN is evaluated on three standard datasets: Iris Flower, Breast Cancer Wisconsin, and Wine. The results presented in this paper are derived from 100 epochs using 10-fold cross-validation, ensuring that all dataset samples are involved in both the training and validation stages. Moreover, the results show that FOXANN outperforms traditional ANN and logistic regression methods as well as other models proposed in the literature such as ABC-ANN, ABC-MNN, CROANN, and PSO-DNN, achieving a higher accuracy of 0.9969 and a lower validation loss of 0.0028. These results demonstrate that FOXANN is more effective than traditional methods and other proposed models across standard datasets. Thus, FOXANN effectively addresses the challenges in ML algorithms and improves classification performance.

## 1. Introduction

Machine learning (ML) consist of algorithms and techniques that aim to train machines (computers) to solve real-world problems, increase performance, increase production, enhance efficiency, and reduce errors caused by humans or traditional techniques [1]. Supervised learning is a subset of the ML field. It is termed *supervised* because the machine is provided with the inputs and outputs (target) during the training process. This type of training lets the algorithm know the relationships between the inputs and output to achieve optimal results in the prediction process when it has been fed with inputs only [2].

Artificial neural networks (ANNs) are considered the most used supervised classifiers and have solved problems effectively since they were developed [3]. Furthermore, ANNs simulate the human brain's work by modeling it mathematically; for example, it mimics the connections among neurons, the functions of neurons, and the memory of the neurons [4], [5]. Research is continuing to improve the field of machine learning in general and neural networks in particular because ANNs play an essential role in most machine learning algorithms and problems such as convolutional neural networks (CNNs), deep learning, transfer learning, natural language processing (NLP), and generative adversarial network (GAN) [6], [7].

Improving the neural network is necessary due to increases in data and its complexity, computer resources, cloud solutions, cybersecurity threats, and generative AI [8]. However, optimization is widely applied in ANN such as swarm algorithms or evolutionary algorithms, i.e., artificial bee colony (ABC), ant colony optimization (ACO), particle swarm optimization (PSO), genetic algorithms (GA), and differential evolution (EA). Moreover, some researchers have examined the tuning of ANN hyperparameters, such as learning rate, initial weight, momentum, and epochs, while others have attempted to choose the best structure for an ANN (i.e., the number of hidden layers and the size of the hidden layer) [9], [10].

This paper proposes an improved ANN based on the Fox optimizer (FOX), a state-of-the-art metaheuristic optimization algorithm that simulates the behavior of red foxes hunting prey [11]. This novel classification model, called FOXANN, aimed to increase accuracy and reduce loss in classification problems by eliminating the backpropagation algorithm from the ANN and employing FOX to optimize the synaptic weights. Moreover, the authors utilized FOX because it uses a static balance between exploitation and exploration that leverages the diversity in the search space to reduce the probability of entrapment in local optima. In contrast, the backpropagation algorithm uses gradient descent methods that focus on exploitation only, enabling the process of training ANN entrapped with local optima, especially when the number of samples is insufficient.

Three standard datasets have been used to evaluate the proposed FOXANN. First, the Iris Flower dataset consists of 150 instances and contains four features, with three classes: Setosa, Versicolor, and Virginica. Second, the Wine dataset consists of 178 instances and 13 features, with three classes determining the type of Wine. Lastly, Breast Cancer Wisconsin dataset consists of 569 instances and 30 features with two classes: Malignant and Benign [12].

The main contributions of this paper are highlighted as follows:
1- FOXANN utilizes the FOX algorithm instead of the backpropagation algorithm in an ANN to avoid the problem of local optima.
2- This replacement improves accuracy, robustness against local optima, and interpretability due to FOX's superior optimization capabilities compared to other optimization techniques.
3- FOXANN provides a deeper understanding of decision-making processes by optimizing synaptic weights more effectively than other techniques, and thus, FOXANN represents a significant advancement in the field of ML algorithms.

The rest of this paper is organized as follows: Section 2 presents the latest studies that have used optimization methods to improve ANNs. Section 3 presents the materials and methods used in this paper. Sections 4 and 5 list and discuss the results with comparisons. Finally, Section 6 presents conclusions and suggestions for future directions in ANN improvement.

## 2. Related work

This section reviews the related literature on ANN improvement using different optimization techniques. In recent years, many optimizers, such as ABC, GA, EA, and PSO, have been used to improve ANN, and these improvements have been crucial for solving ML problems.

The study by [13] proposed a hybrid model that utilized ABC and an ANN to efficiently optimize a set of neuron connection weights for an ANN, producing an alternative scheme to traditional ANN training used for short-term electric load prediction. The results of the proposed model showed an improved version of the ANN with higher accuracy and a faster convergence rate with prediction tasks [13]. A recently proposed model that adopts ABC used modular neural networks (MNNs) to optimize the weights. The results of ABC-MNN showed generalization performance with higher training accuracy on the Iris Flower dataset [14]. Moreover, a model called CROANN, which uses chemical reaction optimization (CRO), was proposed to design the structure and tune the weights of an ANN. CROANN was evaluated using the Iris Flower and Breast Cancer Wisconsin datasets, and the results showed that CRO is superior to many EA strategies with ANN [15]. Furthermore, six hybrid ANN models based on adaptive EA were proposed to optimize ANN parameters and feature selection evaluated on two meteorological datasets as prediction tasks. The results showed that these proposed models can be used as generic models with improved forecasting accuracy [16]. A multi-objective optimizer called GA-ANN-GA, was presented by using GA before training to tune the weight of an ANN automatically. Then, the GA was used again after the ANN training phase; this model was used to optimize heat transfer in rectangular perforated plate fins and the results show that optimized fin geometry can be used in many heat transfer problems with less effort [17]. Another novel hybrid algorithm called BRKGA-NN that used GA with ANN was proposed to determine the connection weights, bias values of the hidden neurons, and the number of hidden neurons in an ANN. The BRKGA-NN was evaluated on time-series datasets, and the results showed that BRKGA-NN provided more effective predictions than support vector regression (SVR), BPANN, and autoregressive integrated moving average (ARIMA) [18]. PSO was integrated with deep neural networks (DNNs) in a recent study [19] that proposed a new method called PSO-DNN to optimize the number of hidden layer nodes (neurons). An evaluation of PSO-DNN on digital modulation classification tasks confirmed that the proposed method is effective in optimizing DNN [19], [20]. A recent study suggested a CNN hyperparameter optimization method using linearly decreasing weight PSO (LDWPSO). The proposed LDWPSO-CNN method was evaluated on MNIST and CIFAR-10 datasets. The results showed increased accuracy by 4% from a baseline CNN on the MNIST dataset and by 41% on the CIFAR-10 dataset for 10 epochs [21], [22].

**Table 1:** *Summary of hybrid ANN models based on optimization techniques*

| Study | Model | Dataset | Task | Results |
|---|---|---|---|---|
| [13] | ABC-ANN | Electric Load | Prediction | Improved ANN accuracy and faster convergence rate. |
| [14] | ABC-MNN | Iris Flower | Classification | Higher training accuracy with generalization performance. |
| [15] | CROANN | Iris Flower, Breast Cancer Wisconsin | Classification | Superior performance compared to many EA strategies with ANN. |
| [16] | Adaptive EA-ANN | Meteorological | Prediction | Improved prediction accuracy. |
| [17] | GA-ANN-GA | Heat Transfer | Optimization | Optimized fin geometry with less effort for heat transfer problems. |
| [18] | BRKGA-NN | Time-Series | Prediction | More effective predictions compared to SVR, BPANN, and ARIMA. |
| [19] | PSO-DNN | Digital Modulation | Classification | Effective optimization of DNN for digital modulation classification tasks. |
| [22] | LDWPSO-CNN | MNIST, CIFAR-10 | Classification | Increased accuracy compared to CNN on MNIST and CIFAR-10 datasets. |

Table 1 summarizes related work, including the proposed models, datasets, evaluation tasks, and key findings. Furthermore, several studies have contributed to improving ANNs based on various optimization methods, which were reviewed in this paper [23].

## 3. Materials and methods

This section explains the proposed method by presenting a detailed overview of the working principles and definitions of ANN, FOX, and the proposed FOXANN.

### 3.1. Artificial neural networks

The ANN consists of three main layers: the input, hidden, and output layers; each layer is a cluster of multiple nodes (neurons). The input layer accepts and feeds the input data to the hidden layer. Then, the output layer computes the last result (see Figure 1) [24]. The direction of data from the input to the output is called a feed-forward neural network. ANNs can be used for several tasks, including classification, clustering, and pattern recognition. The input data are multiplied by the value of the weights and passed to the hidden layer neurons, each of which has an activation function to compute the neuron's output as seen in Equation 1 [25]:

$$z_j = f\left(\sum_{i=1}^{n}(x_i \times w_{ij}) + b_j\right) \tag{1}$$

and

$$f = \frac{1}{1+e^{-x}} \tag{2}$$

where $x_i$ is the data input, $w_{ij}$ is the weight between the $i^{th}$ input neuron and the $j^{th}$ hidden neuron, $b_i$ is the bias of the $j^{th}$ hidden neuron, and $z_j$ represents the output of the $j^{th}$ neuron in the hidden layer. Moreover, the neuron's output ($z_j$) is fed into f activation function (i.e., sigmoid Equation. 2). Many activation functions exist, such as sigmoid, tanh, RELU, and softmax. Each function is problem-dependent (i.e., the sigmoid function, Equation 2, is employed when the output range is expected to be between zero and one, in contrast, the tanh function is employed when the output is between -1 and 1) [26].

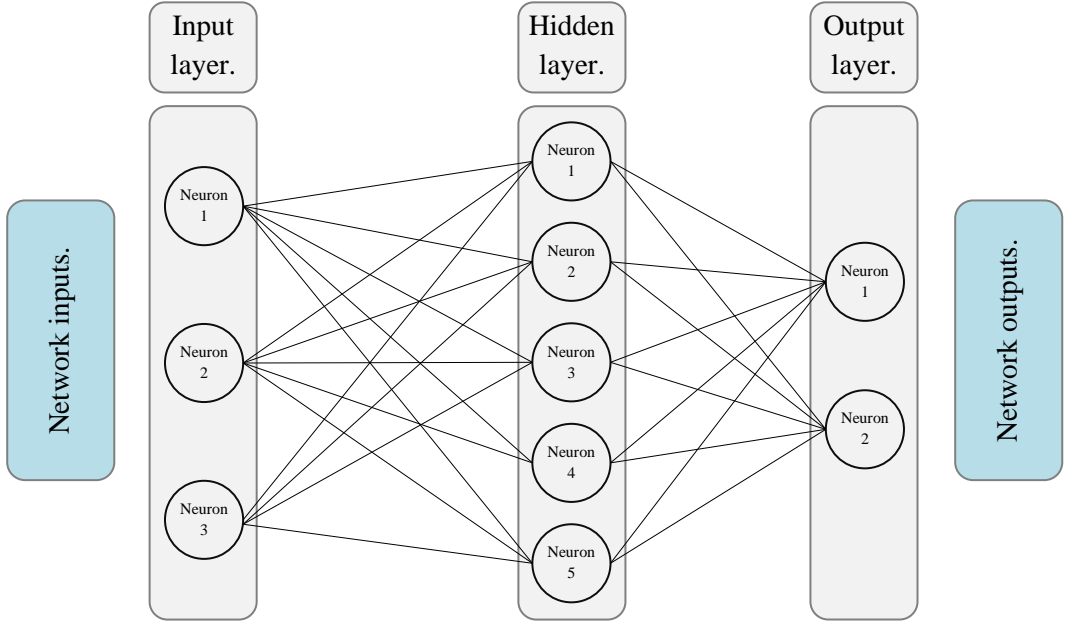

**Figure 1:** Artificial neural network architecture [27]

The input of ANN must be standardized to reduce the variance between data. Moreover, if the output is text data, it must be converted into numerical form to be accepted by the neural network functions. ANNs require standardized input to mitigate the variance between data and provide optimal performance. Additionally, if the output (label) is text data, it must be converted into a numerical format compatible with the neural network's functions. This conversion and standardization enable the network to process and analyze the data effectively [28].

$$E = \frac{1}{2}\sum_{i=1}^{n}(y_i - y'_i)^2 \quad (3)$$

$$\delta_j = (y_j - y'_j) \cdot f'(z_j) \quad (4)$$

$$w_{ij} = w_{ij} + \eta \cdot x_i \cdot \delta_j \quad (5)$$

In the above equations, E is the error between the expected ($y_i$) and actual ($y'_i$) outputs, $\delta_j$ is the error gradient for the unit $y_j$, $f'(z_j)$ is the derived activation function, $w_{ij}$ is the weight connecting unit i to j, updated with learning rate η and input $x_i$ [29], [30].

The most important algorithm in an ANN is backpropagation, which involves feeding the output to the network input again to update the weights. Backpropagation has four main steps: the first is the computation of the error between the expected and the actual output (target) using error functions like mean squared error (MSE), as seen in Equation 3. The second is deriving the error gradient from outputs, which are used to update the weights for the last layer in the network using Equation 4. The third is error backpropagation, in which the error propagates to other layers. Lastly, the weights are updated based on the derived error and learning rate by Equation 5 [31].

### 3.2. Fox Optimizer Algorithm

FOX is an optimization method motivated by red foxes' hunting behavior. It searches for the best position (solution) using static exploration and exploitation. Through exploration, FOX uses a random walk policy, aided by its ability to detect ultrasound to find prey. Upon detecting prey, the FOX agent enters the exploitation phase and waits for the sound of the prey's ultrasound. The FOX agent evaluates the time required to catch its prey based on the sound's travel time and then jumps [32]. The FOX algorithm requires two inputs: an objective function and bounds. The objective function calculates the fitness value, while the bounds determine the range of values for each variable in the optimization problem. Algorithm 1 below introduces a visual representation of how the FOX is used to find the best position; further details can be found in [11].

---

**Algorithm 1:** FOX optimizer [11]
**Input:** Objective function, Problem bounds

1: Initialize the red fox population (X)
2: **While iter < Max_iter**
3:       Initialize variables
4:       Calculate the fitness of each search agent
5:       Select **Best_Position** and **Best_Fitness** among the population
6:       **If** fitness$_{i+1}$ < fitness$_i$
7:             Best_Fitness = fitness$_{i+1}$
8:             Best_Position = X$_{(i)}$
9:       **EndIf**
10:     **If** r >= 0.5
11:           Initialize time randomly;
12:           Calculate Distance_Sound_travels
13:           Calculate Sp_S
14:           Calculate the distance from fox to prey
15:           Tt = average time;
16:           T = Tt/2;
17:           Calculate jump
18:           **If** p > 0.18
19:                 Find X$_{(i+1)}$ using Equation 6
20:           **Elseif** p <= 0.18
21:                 Find X$_{(i+1)}$ using Equation 7
22:           **EndIf**
23:     else
24:           Find MinT
25:           Explore X$_{(i+1)}$ using Equation 8
26:     **EndIf**
27:     Clip the position if it goes beyond the limits
28:     Evaluate search agents by fitness
29:     Update Best_Position
30:     iter = iter + 1
31: **EndWhile**
32: **return** Best_Position & Best_Fitness

**Output:** Best Solution, Best Fitness

---

FOX employs a static trade-off between exploration and exploitation (50% for each). The algorithm uses random walks to find the red fox's prey during exploration. During exploitation, the algorithm calculates the distance to the prey, jump height, and new positions.

$$X_{(i+1)} = DFP_i * Jump_i * c_1 \qquad (6)$$

$$X_{(i+1)} = DFP_i * Jump_i * c_2 \qquad (7)$$

The two constants, $c_1$ and $c_2$, have been fixed at 0.18 and 0.82, respectively. These constants are derived from observations made while studying the jumping behavior of a red FOX. It has been observed that the FOX's jumps are either directed toward the northeast or the opposite direction. The FOX explores the Red Fox's surroundings by using the following equation to calculate its new position (this is considered exploration) [33]:

$$X_{(i+1)} = BestPosition * rand(1, dimension) * MinT * a \qquad (8)$$

and

$$tt = \frac{sum(Time_{st_{it}})}{dimension}, \quad MinT = Min(tt) \qquad (9)$$

$$a = 2\left(iter - \frac{1}{max(iter)}\right) \qquad (10)$$

where tt is the time average equal to the summation of the time variable divided by the dimension of the problem, $iter$ is the current iteration, and $max(iter)$ is maximum iterations. Calculating both the $MinT$ and $a$ variable has a vital effect on the search phase to move toward a solution that is close to the best solution. Using a random function $rand(1, dimension)$ ensures that the fox walks stochastically to explore the prey [11], [34].

### 3.3. Fox Artificial Neural Network

The proposed FOXANN method is inspired by the use of FOX to tune the hyperparameter automatically in Q-learning [35] and by FOX's superior performance in solving common optimization problems. This section presents an AAN based on FOX intended to solve classification tasks in standard datasets, which are expected to improve the ANN's ability to solve ML problems. The ANN structure in the proposed method, FOXANN, remains unchanged, and the essential improvement lies in the backpropagation algorithm. Since this step was eliminated, the authors use FOX to improve the weights based on the minimization of MSE in Equation 3.

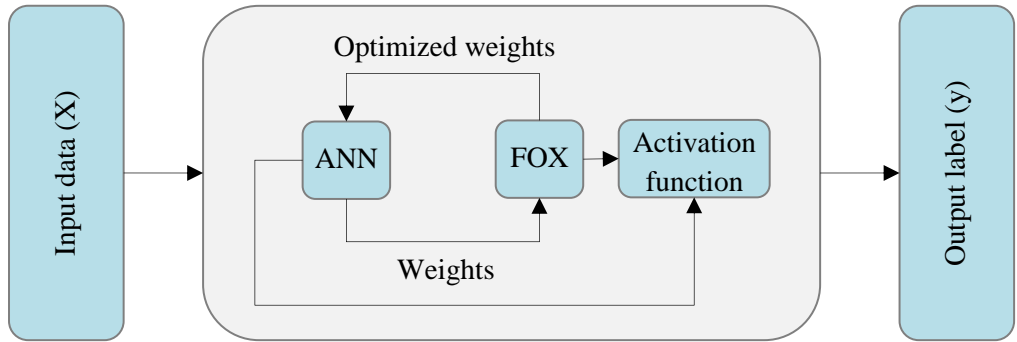

**Figure 2:** FOXANN architecture.

The proposed model's architecture is simple, as shown in Figure 2. Once the data have been processed for input, they are moved directly into the ANN. The feed-forward process is then performed, where FOX iteratively optimizes the weights with minimum loss and returns them to the ANN. The ANN uses the optimized weights and activation function to compute the final output based on Equation 1.

### 3.4. Materials

The Iris Flower, a standard benchmark dataset in ML, contains 150 objects and four features. These features represent the sepals and petals of the iris flower, with three distinct classes: Setosa, Versicolor, and Virginica [36]. Furthermore, a well-known Breast Cancer Wisconsin dataset offers insights into breast cancer diagnosis with 569 instances and 30 features. Features are extracted from fine needle aspirates of breast tissue. This dataset provides valuable information for classifying malignant and benign tumors [37]. Moreover, the Wine dataset, which contains 178 instances and 13 features, including attributes such as alcohol content, acidity levels, and color intensity, is used to classify wines into one of three classes [21]. These datasets are explored in Table 2 and are basic for classification tasks, helping researchers to explore and evaluate various ML algorithms.

**Table 2:** *Datasets' characteristics*

| Dataset | Instances | Features | Classes |
|---|---|---|---|
| Iris Flower | 150 | 4 | 3 |
| Breast Cancer Wisconsin | 569 | 30 | 2 |
| Wine | 178 | 13 | 3 |

All the datasets have been preprocessed using min-max normalization from Equation 11 to ensure that different feature scales are processed in the same range, typically between 0 and 1. Moreover, the preprocessing using min-max normalization helps to reduce variations in feature scales, ensuring that the model remains robust against outliers.

$$X_{normalized} = \frac{X - X_{min}}{X_{max} - X_{min}} \quad (11)$$

In the above equation, X is the feature vector used by the models as input, $X_{min}$ is the minimum value in the vector, and $X_{max}$ is the maximum value in the vector.

## 4. Results

The performance of the FOXANN model was evaluated using the Iris Flower, Wine, and Breast Cancer Wisconsin datasets, considering standard evaluation metrics such as validation loss, accuracy, recall, precision, and F-score. The ANN topology has been designed based on a previous study [38] that recommends using two hidden layers with a small-to-median dataset. The first layer was twice the size of the input layer, and the second layer was half the size of the input layer to better generalization. Additionally, the initial weights of the ANN range [-3, 3] are based on a trial-and-error strategy. In this paper, many experiments have been conducted to evaluate the proposed model's performance against traditional ANN and logistic regression (LR) classifiers. Furthermore, it compares the most important results from the literature with the proposed model FOXANN. The figures and tables below show the results obtained after 100 epochs using the cross-validation method with 10 folds, which ensures that all data are involved in the training process for each classifier [39]. In ML, visualizing the results is a crucial step when analyzing and comparing the performances of different methods. The performance of each classifier's training process was measured by computing the validation loss using the MSE. Figures 3 (A), (B), and (C) show the validation loss on the Iris Follower dataset, Breast Cancer Wisconsin dataset, and Wine dataset, respectively. Furthermore, Figure 3 (D) presents the average validation loss for all datasets.

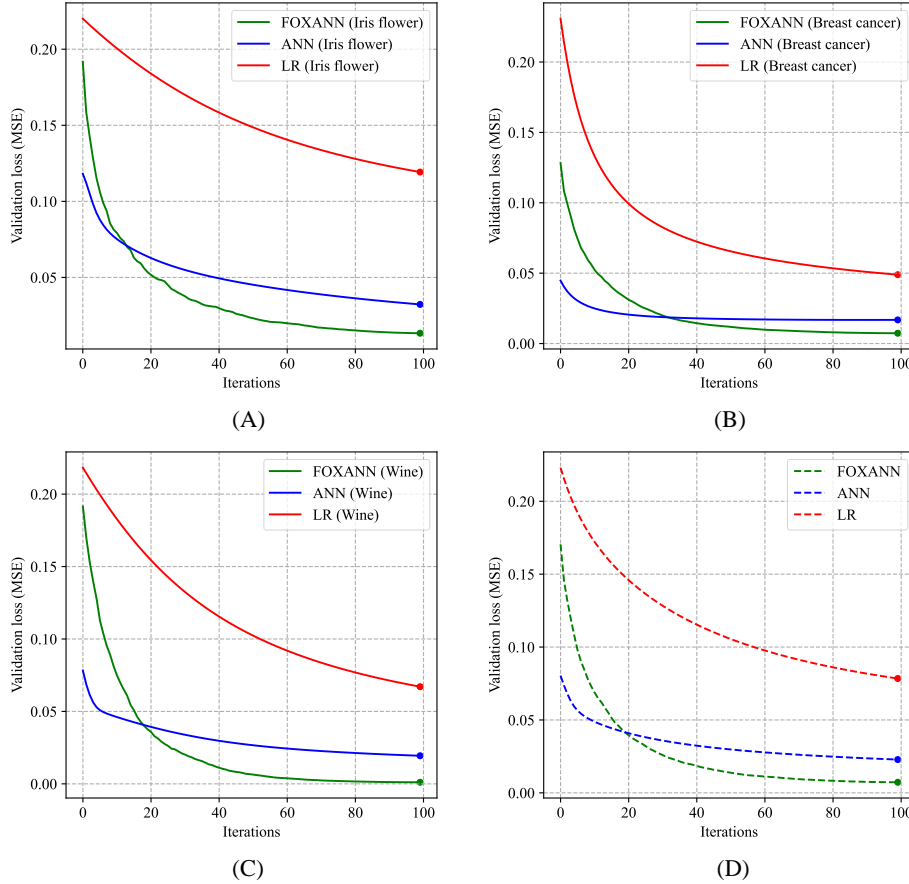

**Figure 3:** Validation loss on the dataset: (A) Iris Flower, (B) Breast Cancer Wisconsin, (C) Wine, and (D) average validation loss.

Table 3 shows the results of the evaluations performed on three different datasets using the proposed FOXANN, ANN, and LR. Firstly, the Iris Flower dataset was evaluated using FOXANN and reached an accuracy of 0.9776 and a validation loss of 0.0107. The ANN achieved an accuracy of 0.9754 and a validation loss of 0.0296, and the LR produced an accuracy of 0.8555 with a validation loss of 0.1193. The Breast Cancer Wisconsin dataset was then evaluated using FOXAN and achieved an accuracy of 0.9749 with a validation loss of 0.0075, while the ANN achieved an accuracy of 0.9654 with a validation loss of 0.0201, and LR produced an accuracy of 0.9596 with a validation loss of 0.0489. Lastly, the evaluation of the Wine dataset using FOXANN achieved an accuracy of 0.9969 with a validation loss of 0.0028, while the ANN achieved an accuracy of 0.9844 with a validation loss of 0.0137, and LR produced an accuracy of 0.9421 with a validation loss of 0.0671.

**Table 3:** *FOXANN Performance results*

| Dataset | Model | Accuracy | Loss | Precision | Recall | F-Score |
|---|---|---|---|---|---|---|
| Iris Flower | FOXANN | 0.9776 | 0.0107 | 0.9664 | 0.9664 | 0.9664 |
| | ANN | 0.9754 | 0.0296 | 0.9664 | 0.9598 | 0.9631 |
| | LR | 0.8555 | 0.1193 | 0.9891 | 0.5733 | 0.7239 |
| Breast Cancer Wisconsin | FOXANN | 0.9749 | 0.0075 | 0.9749 | 0.9749 | 0.9749 |
| | ANN | 0.9654 | 0.0201 | 0.9654 | 0.9654 | 0.9654 |
| | LR | 0.9596 | 0.0489 | 0.9596 | 0.9596 | 0.9596 |
| Wine | FOXANN | 0.9969 | 0.0028 | 0.9834 | 0.9875 | 0.9902 |
|  | ANN | 0.9844 | 0.0137 | 0.9844 | 0.9788 | 0.9816 |
|  | LR | 0.9421 | 0.0671 | 0.9684 | 0.854 | 0.9076 |
| **Average results** | **FOXANN** | **0.9831** | **0.0070** | **0.9749** | **0.9763** | **0.9772** |
|  | ANN | 0.9751 | 0.0211 | 0.9721 | 0.968 | 0.9700 |
|  | LR | 0.9191 | 0.0784 | 0.9724 | 0.7956 | 0.8637 |

Furthermore, Table 3 shows that the average accuracy and average validation loss using FOXAN, ANN, and LR, were 0.9831, 0.9751, and 0.9191, respectively. It also presents important metrics, such as precision, recall, and F-score.

## 5. Discussion

The results in Table 3 and Figure 3 show that FOXANN performs well compared with ANN and RL methods, and it outperformed both the ANN and LR for the Iris Flower dataset, achieving the highest accuracy of 0.9776. This indicates that FOXANN effectively learned the patterns in the datasets and produced accurate predictions. It also achieved the lowest validation loss of 0.0107, indicating better generalization ability compared to ANN and LR. In the Breast Cancer Wisconsin dataset, FOXANN performed effectively, with an accuracy of 0.9749 and the lowest validation loss of 0.0075. These results suggest that FOXANN could recognize the fundamental patterns within the datasets effectively, leading to accurate predictions and performance. The ANN performed well, with an accuracy of 0.9654 and a validation loss of 0.0201, and LR performed the worst, with an accuracy of 0.9596 and the highest validation loss of 0.0489.

The FOXANN performed better than the ANN and LR in evaluating the Wine dataset, achieving the highest accuracy of 0.9969 and the lowest validation loss of 0.0028. This result shows that FOXANN can effectively identify the patterns in the Wine dataset, resulting in superior performance. The average accuracies for FOXANN, ANN, and LR were 0.9831, 0.9751, and 0.9191, respectively; the validation losses were 0.007, 0.0211, and 0.0784, respectively.

**Table 4:** *Comparison between FOXANN and other models*

| Model | Reference | Dataset | Accuracy |
|---|---|---|---|
| **FOXANN** | Proposed | Iris Flower | **0.9776** |
| ABC-NN | [14] |  | 0.9666 |
| ABC-MNN | [14] |  | 0.9722 |
| CROANN | [15] |  | 0.9656 |
| **FOXANN** | Proposed | Breast Cancer Wisconsin | **0.9749** |
| Decision tree (DT) | [20] |  | 0.9400 |
| DT (PCA Transformation) | [20] |  | 0.9600 |
| CROANN | [15] |  | 0.9611 |
| **FOXANN** | Proposed | Wine | **0.9969** |
| RFC | [21] |  | 0.8979 |
| KNN | [21] |  | 0.8666 |

Finally, FOXANN surpassed both the ANN and LR algorithms due to its unique capabilities based on the static balance between exploration and exploitation, which effectively mitigates the local optima trap in the backpropagation algorithm. Table 4 highlights FOXANN's dominance over the proposed models in the literature. It outperformed the ANN-based optimizers, such as ABC, CRO, and RFC, in terms of accuracy. Additionally, FOXANN surpassed GA, PSO, EA, and LDWPSO in terms of interpretability and generality, with a significantly low error rate of 0.007 compared to 0.01 (GA), 0.009 (PSO and LDWPSO), and 0.022 (EA). These results emphasize FOXANN's broader applicability and superior interpretability compared to ML methods that operate based on best optimizers.

## 6. Conclusion

This paper presents a novel model called FOXANN, which combines the FOX optimization algorithm with an ANN algorithm to effectively increase performance in solving machine learning problems. The backpropagation algorithm is replaced with FOX to optimize the weights, improve classification performance, and avoid the local optima trap that may be caused by the backpropagation algorithm. Experimental results on standard datasets (Iris Flower, Breast Cancer Wisconsin, and Wine) show that FOXANN achieves higher accuracy and lower validation loss than traditional ANN and LR methods, as well as other methods proposed in the literature. Future studies may focus on enhancing the FOX optimizer and integrating it with more complex ML models, such as deep learning models, while considering imbalanced or imagery datasets. Furthermore, the FOX algorithm might be used to present optimal ANN or CNN structure designs based on the problem features to reduce the model's complexity.